%% file: main.tex
\documentclass[runningheads]{llncs}

\usepackage{eccv}

\usepackage{eccvabbrv}

\usepackage{graphicx}
\usepackage{booktabs}

\usepackage[accsupp]{axessibility}  

\usepackage{algorithm}
\usepackage{algpseudocode}

\usepackage{multirow}
\usepackage{diagbox}
\usepackage[table]{xcolor}
\usepackage{lipsum}
\usepackage{tabularray}

\usepackage{hyperref}

\usepackage{orcidlink}

\begin{document}

\title{MaCo-GAN: Manifold-Contrastive Adversarial Learning for Single Image Super-Resolution} 

\titlerunning{MaCo-GAN}

\author{Daeyoung Han\inst{1}\orcidlink{0000-0003-0368-8675} \and
Seongmin Hwang\inst{2}\orcidlink{0000-0003-3313-8586} \and
Moongu Jeon\inst{1}\orcidlink{0000-0002-2775-7789}}

\authorrunning{D. Han, et al.}

\institute{
Department of Electrical Engineering and Computer Science, Gwangju Institute of Science and Technology, Gwangju, Republic of Korea \\
\email{\{xesta120, mgjeon\}@gist.ac.kr} \and
Department of AI Convergence, Gwangju Institute of Science and Technology, Gwangju, Republic of Korea \\
\email{sm.hwang@gm.gist.ac.kr}
}

\maketitle

\input{contents/abstract}
\input{contents/intro}
\input{contents/related_works}
\input{contents/proposed_methods}
\input{contents/experiments}
\input{contents/conclusion}

\bibliographystyle{splncs04}
\bibliography{main}

\include{contents/supplementary}

\end{document}

%% file: contents/abstract.tex
\begin{abstract}
Conventional Generative Adversarial Networks (GANs) for Single Image Super-Resolution (SISR) often struggle with hallucinated artifacts, largely because standard discriminators evaluate overall image naturalness rather than strict conditional realism. 
To address this, we propose MaCo-GAN, a novel manifold-contrastive GAN framework that replaces the conventional adversarial loss with a supervised contrastive objective. 
A core component of our method is a dynamic fake sample synthesizer that transforms ground truth (GT) data into a spectrum of challenging, perceptually plausible fake images that strictly maintain low-resolution (LR) correspondence. 
Utilizing these synthesized samples, we establish a robust contrastive minimax game: the generator is trained to attract its predictions toward \textit{on-manifold} fakes (low distortion) and repel them from \textit{off-manifold} fakes (high distortion), while the discriminator optimizes the exact opposite. 
By simply replacing the adversarial loss of a baseline SR model with our proposed objective, we demonstrate consistent improvements in the perception-distortion trade-off across various benchmarks. 
Extensive ablation studies validate the effectiveness of our framework and provide deep insights into the dynamics of this conditional contrastive game.
\keywords{Single Image Super-Resolution \and Contrastive Learning \and Generative Adversarial Networks}
\end{abstract}

%% file: contents/intro.tex
\section{Introduction}
\label{sec:intro}

Single Image Super-Resolution (SISR) is a fundamental computer vision task that aims to reconstruct a high-resolution (HR) image from a single low-resolution (LR) counterpart. 
This problem is inherently ill-posed, as a single LR image can correspond to multiple potential HR images.
Conventional learning-based SISR methods have historically approached this by minimizing a pixel-wise loss (e.g., L1 or L2) between the prediction and the ground-truth (GT) image. 
However, as is widely known~\cite{ledig2017photo}, this objective function leads to the ``blurring phenomenon'', by forcing the model to find the averages of plausible solutions, thereby collapsing perceptual variance~\cite{lee2025auto, james2003variance, lee2019harmonizing}.

To overcome this limitation, Generative Adversarial Networks (GANs)~\cite{goodfellow2014generative} were introduced to the field of perceptual SISR.~\cite{ledig2017photo}
By bringing a generator (an SISR network) and a discriminator into a minimax game, models in this field optimize perceptual quality, producing images that appear more natural and realistic. 
In standard GANs, the discriminator estimates the probability that the input image is real.
It learns to distinguish between real and generated data by minimizing the probability that the SISR prediction is real, while its opponent attempts to deceive it by maximizing that probability.
Subsequent work on relativistic GANs (RAGAN)~\cite{jolicoeur2018relativistic} improved this framework by introducing real data into the generator's loss as well.
It benefits from the signals of both real and generated samples, outperforming the original GAN in terms of stability and perceptual quality.~\cite{wang2018esrgan}

\begin{figure}[t]
    \centering
    \includegraphics[width=1\linewidth]{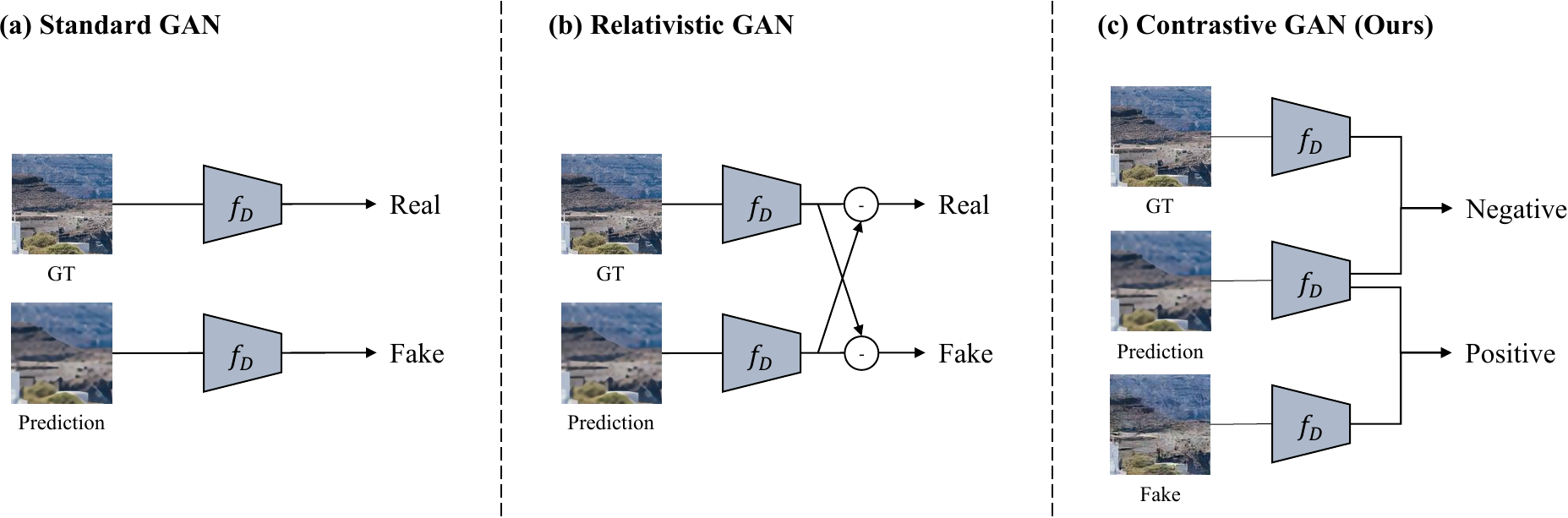}
    \caption{
        Graphical comparison of our proposed contrastive GAN with existing frameworks in SISR.
    }
    \label{fig:intro}
\end{figure}

Despite its excellence, training GANs for SISR remains challenging. 
The adversarial game is still unstable, and the discriminator can often overpower the generator, causing the training to collapse and fail to reach an equilibrium. 
In this paper, we enhance this GAN-based framework by proposing \textit{MaCo-GAN} (Manifold-Contrastive GAN), a novel contrastive strategy described in \cref{fig:intro}. 
Our core idea is to replace the conventional adversarial loss, which only compares outputs with real data, with a supervised contrastive objective~\cite{khosla2020supervised}. 
Instead of a simple one-to-one comparison with the GT, our framework teaches the SISR network to navigate the solution space more accurately and effectively. 
It learns to pull its predictions toward a set of desirable solutions and push them from a set of undesirable solutions, all within the bounds of solutions that can be degraded to a given LR.

The success of this strategy hinges entirely on the quality of these synthetic samples.
We introduce a learning-based strategy, the \textit{fake sample synthesizer}, which serves as the core component of our method. 
This network is trained to randomly generate a diverse spectrum of challenging fakes from the GT image by controlling the level of distortion with a parameter. 
To create meaningful and diverse negative samples, we corrupt a real image with random noise, and the synthesizer reconstructs it into various fake images that are both perceptually natural and highly LR-corresponding, serving as challenging negative targets for the SISR network.

We utilize this synthesizer to establish a novel contrastive minimax game. 
For a given query (the SR prediction), we define the GT or marginally corrupted fake samples as positive samples of the SISR network.
Highly distorted fakes are then assigned as negatives, representing what the SISR network desires its output not to be.
While the SISR network learns to minimize the supervised contrastive loss with these key sets, the discriminator is trained with the exact opposite setting. 

In this work, we demonstrate that such a ``many-to-many'' comparison provides a more beneficial signal to perceptual SISR than traditional GAN losses. 
Our contributions are as follows:
\begin{itemize}
    \item We propose MaCo-GAN, a novel contrastive GAN for perceptual SISR that replaces the standard adversarial loss with a supervised contrastive loss.
    \item We introduce a fake sample synthesizer that creates plausible and LR-corresponding fake samples from real data, representing the solution space of the GT.
    \item We define a contrastive minimax game where the SR network and the discriminator compete to minimize their conflicting contrastive objectives.
    \item Our method's effectiveness is validated by demonstrating improved quantitative performance and visual quality compared to previous methods.
    \item Through extensive analyses, we dissect the dynamics of the contrastive GAN and explore strategies to improve the system by stabilizing the discriminator.
\end{itemize}

%% file: contents/related_works.tex
\section{Related Works}
\label{sec:related_works}

\subsubsection{GAN-based SISR}
To address the limitation that SISR methods that focus on minimizing pixel-wise loss functions tend to produce overly smooth and fail to capture the perceptual quality of natural images, the pioneering SRGAN~\cite{ledig2017photo} introduced an adversarial loss~\cite{goodfellow2014generative} to improve perceptual quality.
From the perspective of conditional image generation, this framework divides the optimization objectives into two parts: the content loss (pixel-wise or VGG-based perceptual) enforces spatial alignment with the corresponding GT, while the adversarial loss encourages overall perceptual realism.
This approach is similar to AC-GAN~\cite{odena2017conditional} that utilizes an auxiliary classifier to align the class of generated samples with GTs.

Following this, several subsequent works~\cite{wang2018esrgan, wang2021real} proposed different types of adversarial losses~\cite{jolicoeur2018relativistic, schonfeld2020u} from the original GAN.
Specifically, the discriminator of ESRGAN assesses how much an image is relatively more realistic than another, while the discriminator of SRGAN just classifies whether an image is real or not.
State-of-the-art studies~\cite{liang2022details, lee2025auto, xu2024uncovering} have shown the effectiveness of this approach in perceptual SISR.

\subsubsection{Contrastive Learning}
In recent years, contrastive learning has emerged as a powerful methodology for representation learning, particularly in the self-supervised domain~\cite{chen2020simple, he2019momentum, chen2020improved, oord2018representation, wu2018unsupervised}. 
The core idea is to learn a rich and meaningful representation space by pulling a query sample closer to positive samples while pushing it away from negative samples.
These works generally leverage the loss function from the noise contrastive estimation~\cite{gutmann2010noise}.

Beyond self-supervised learning, Khosla \etal~\cite{khosla2020supervised} proposed supervised contrastive learning, extending the conventional framework to allow multiple positive samples for a single query. 
While standard contrastive objectives pull a query toward a single positive key, the supervised variant computes the loss over a set of positives, effectively pulling the representation toward a localized cluster of desirable features. 
This ``many-to-many'' comparison provides a highly beneficial signal, particularly for inherently ill-posed tasks like SISR, where a single LR image corresponds to multiple valid HR solutions.

This paradigm has been adapted for SISR by several works. 
Zhang \etal~\cite{zhang2021blind} proposed a two-stage framework for blind SISR, where a feature extractor is optimized via a bidirectional contrastive loss, and a conditional contrastive loss is utilized along with the reconstruction loss.
Wu \etal~\cite{wu2023practical} introduced a supervised contrastive loss that pushes reconstructed images toward sharpened positive samples and repels them from blurred hard negative samples, which are synthesized from the GT.
However, these methods only added their contrastive formulation to the distortion-oriented framework, omitting the perceptual and adversarial loss entirely, which limits their generative capabilities compared to GAN-based frameworks.
In contrast with these non-GAN-based approaches, Xu \etal~\cite{xu2024uncovering} proposed a patch-level pixel-wise contrastive loss
to reduce the entropy and variance of potential high-resolution distributions, attaching it as an auxiliary objective to ESRGAN~\cite{wang2018esrgan}.

In this paper, we suggest a more advanced and efficient approach to applying contrastive learning to perceptual SISR.
While a few works~\cite{cao2017tripletgan, jeong2021training} have introduced similar concepts of replacing the standard adversarial loss with a contrastive objective in image generation, our work is the first to adapt this principle to solve the conditional realism gap in SISR.
Unlike prior SISR methods that rely on expensive auxiliary networks or omit crucial adversarial signals, our proposed MaCo-GAN directly integrates the supervised contrastive objective into the discriminator. 

%% file: contents/proposed_methods.tex
\section{Proposed Methods}
\label{sec:proposed_methods}

\begin{figure*}[t]
    \centering
    \includegraphics[width=1\linewidth]{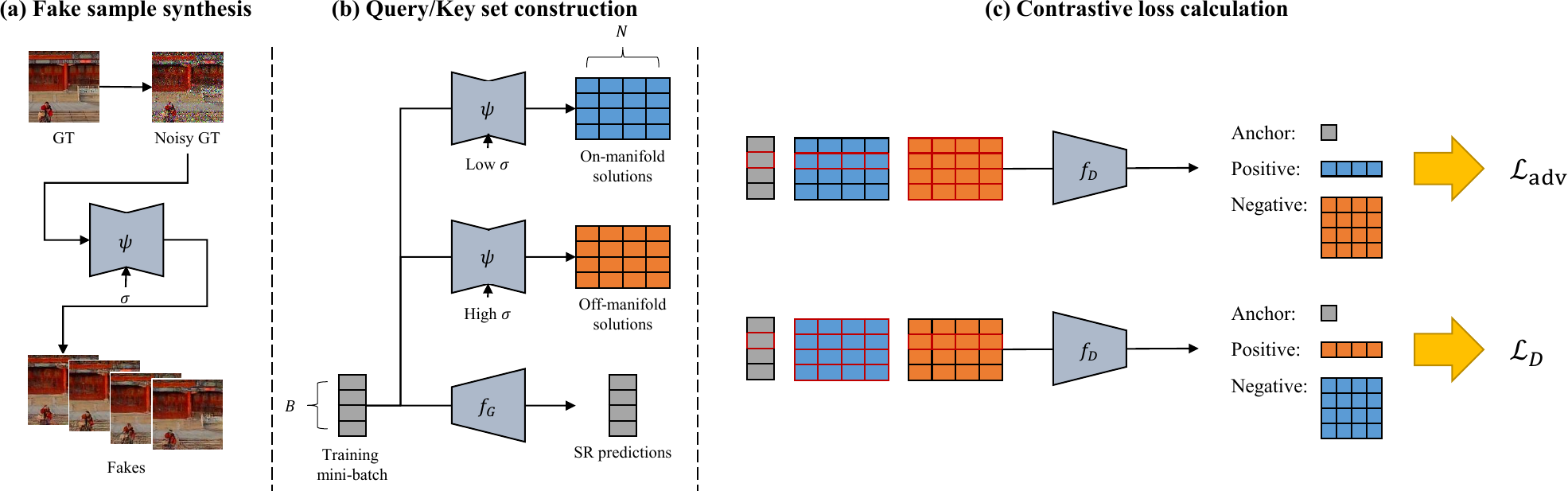}
    \caption{
        Overview of our proposed framework.
        (a) The fake sample synthesizer $\psi(\cdot, \sigma)$ yields random fake samples from the noisy GT, which is corrupted by additive random Gaussian noise with a level of $\sigma$.
        The outputs exhibit diverse high-frequency details while conserving the LR correspondence with the input GT.
        (b) To establish a contrastive learning process, the sets of proper keys for the query (SR prediction) must be constructed.
        In our framework, $B$ GT images in a mini-batch are augmented by performing the fake sample generation $N$ times per image.
        To leverage the supervised contrastive loss, we form two multi-viewed batches whose elements are near or distant from the target manifold, which are built with low and high noise levels, respectively.
        (c) For a minimax game, the generator and discriminator have to aim for minimizing conflicting objective functions: $\mathcal{L}_{\text{adv}}$ and $\mathcal{L}_{D}$, respectively.
        Each loss is calculated with opposite positive/negative setups, which induces an equilibrium of the adversarial game.
    }
    \label{fig:overview}
\end{figure*}



In this section, we propose a simple and novel strategy that uses synthetic samples as negative targets in GAN training.
First, we introduce how to synthesize a set of challenging fake samples from real data.
Second, we suggest a contrastive framework for GAN training that employs these fakes within a contrastive learning objective to refine the SISR model. 
The overview of our proposed method is described in \cref{fig:overview}.

\subsection{Preliminaries}
\label{subsec:preliminaries}

We begin from the common definition of the SISR problem discussed in several previous works~\cite{xu2024uncovering, menon2020pulse, soh2019natural, wang2020deep}.
An LR image $\mathbf{x}_{\text{LR}} \in \mathbb{R}^{3 \times h \times w}$ is obtained by degrading a ground-truth (GT) HR image $\mathbf{x}_{\text{GT}} \in \mathbb{R}^{3 \times H \times W}$:
\begin{equation}
    \mathbf{x}_{\text{LR}} = \mathcal{D}(\mathbf{x}_{\text{GT}}; s),
\end{equation}
where we assume that $\mathcal{D}$ is a known downsampling operation and $s$ is the scale factor.
The SISR problem, which aims to recover the GT from the LR, has an inherent ill-posed nature: multiple HR images can correspond to a single LR image.
We define the set of these HR images as $\mathcal{R} \subset \mathbb{R}^{3 \times H \times W}$, and call it the \textit{feasible set}.
The ultimate goal of SISR is to seek ones that look natural and realistic among them.
In other words, if we take the natural image manifold $\mathcal{M} \subset \mathbb{R}^{3 \times H \times W}$ into account, plausible solutions are on the intersection $\mathcal{M} \cap \mathcal{R}$.

Conventional SISR works have attempted to resolve this ill-posed problem by minimizing a pixel-wise loss between the prediction and the GT.
However, in the context of the bias-variance decomposition~\cite{james2003variance, lee2019harmonizing}, this objective forces the model to reduce both the bias error and the prediction's variance, which results in the widely-known blurring phenomenon~\cite{ledig2017photo}.
The authors of \cite{lee2025auto} term this bias, or the centroid of the plausible solution manifold, the fidelity bias.
Since minimizing $\mathcal{L}_{pix}$ explicitly collapses the perceptual variance, the output of an SR model $f_{\mu}(\cdot)$ pre-trained only using this loss can be considered a strong approximation of this fidelity bias.
In this paper, we leverage this insight and use $f_{\mu}(\mathbf{x}_{\text{LR}})$ to represent the average point of all plausible solutions for a given $\mathbf{x}_{\text{LR}}$.

\subsection{Fake Sample Synthesis}
\label{subsec:fake_sample_synthesis}

Most SISR studies have concentrated only on the plausible solutions satisfying two main conditions: LR correspondence and naturality.
However, we opt to pay attention to feasible but not realistic images.
In the literature of diffusion models, sampling guidances~\cite{dhariwal2021diffusion, ho2022classifier, ahn2024self, hong2023improving} are powerful and effective methods for better performance, which guide outputs to be desirable by pushing them from undesirable samples.
Inspired by these works, we explore a novel strategy utilizing the undesirable HR images as hints, providing the knowledge of what the desirable is to the SISR model.

To deploy this insight, we first need to define the undesirable HR images.
A similar study~\cite{wu2023practical} uses blurry GT images generated with random parameters as undesirable samples.
This approach is effective because blurry samples can be generated quickly and ensure almost perfect LR correspondence.
However, they look very unnatural and lie in a space that is extremely distant from the natural image manifold.
Thus, they cannot push the SISR model to learn the real high-frequency details that are challenging to model.

Beyond the conventional approach, we aim to synthesize multiple random fake samples $\{\dot{\mathbf{x}}^{i}\}_{i=1}^{N}$ with a neural network $\psi(\cdot, \sigma)$, that takes a real GT image as input and modulates the level of perceptual distance from the GT with a noise strength $\sigma$.
For meaningful guidance in the following step, we must ensure the fake samples belong to the feasible set given $\mathbf{x}_{\text{LR}}$.
Meanwhile, the naturality of the fake samples must be guaranteed to a certain degree.
Thus, the main objectives of $\psi$ are as follows:
(1) The downsampled version of $\mathbf{x}_{\text{GT}}$ and $\dot{\mathbf{x}}^{i}$ are very similar, 
(2) $\dot{\mathbf{x}}^{i}$ looks perceptually good, 
for any $i \in [1, N]$.

To achieve these goals, we reflect them in the loss function and the model design of $\psi$.
First, our loss function is, following most of the existing autoencoders~\cite{esser2021taming, weber2024maskbit, yu2023language}, built with a weighted sum of three terms as follows:
\begin{equation}
\label{eq:loss_ae}
    \mathcal{L}_{\text{AE}} =
    \lambda_{\text{pix}}\cdot \mathcal{L}_{\text{pix}} + \lambda_{\text{LPIPS}}\cdot \mathcal{L}_{\text{LPIPS}} + \lambda_{\text{adv}}\cdot \mathcal{L}_{\text{adv}}.
\end{equation}
$\mathcal{L}_{\text{pix}}$ indicates the mean absolute error (MAE) between the downsampled version of the output and GT, which regularizes the output to satisfy the first condition.
$\mathcal{L}_{\text{LPIPS}}$ denotes the LPIPS distance between them, and $\mathcal{L}_{\text{adv}}$ is the adversarial loss.
These two terms guarantee the second condition.
We implement the network $\psi$ based on the U-Net architecture~\cite{ronneberger2015u}. 
To make it operate adaptively to the noise scale $\sigma$, we leverage the design of the DDPM denoiser~\cite{ho2020denoising}, which conditions all blocks on the timestep by adding the corresponding sinusoidal positional embedding.

However, it cannot generate diverse fake samples with a fixed input $\mathbf{x}_{\text{GT}}$.
For more diversity, the input is pre-processed into $\mathbf{x}_{\sigma}$, resulting from adding Gaussian random noise with variance of $\sigma^2$ to $\mathbf{x}_{\text{GT}}$.
In other words, the noisy input $\mathbf{x}_{\sigma}$ is drawn from a Gaussian distribution $\mathcal{N}(\mathbf{x}_{\text{GT}}, \sigma^2 \mathbf{I})$ so that the network, in practice, plays a role in transforming the Gaussian distribution into the sophisticated distribution of the fake samples.

Meanwhile, the additive noise does not need to be isotropic across all image pixels.
As discussed in \cite{ning2021uncertainty}, high-frequency areas like edges and textures have higher uncertainty compared to smooth areas.
To model this characteristic properly, we assign higher variance to regions where fidelity-oriented SISR models struggle, and vice versa.
With the pre-trained $f_{\mu}$, the uncertainty on each pixel is generally modeled to be proportional to the residual between the GT image and the fidelity bias $\mathbf{x}_{\mu} = f_{\mu}(\mathbf{x}_{\text{LR}})$.
Thus, we construct the noisy input by adding anisotropic Gaussian random noise to the GT image like this:
\begin{equation}
    \mathbf{x}_{\sigma} = \mathbf{x}_{\text{GT}} + \sigma \cdot |\mathbf{x}_{\mu}-\mathbf{x}_{\text{GT}}| \odot \epsilon, \quad \epsilon \sim \mathcal{N}(\mathbf{0}, \mathbf{I}),
\end{equation}
where $\odot$ denotes the Hadamard product.
In this paper, we set the range of noise level to $[0.0, 5.0]$.
The overall synthesis process is described in \cref{alg:algorithm1}.

\begin{figure*}[t]
    \centering
    \includegraphics[width=1\linewidth]{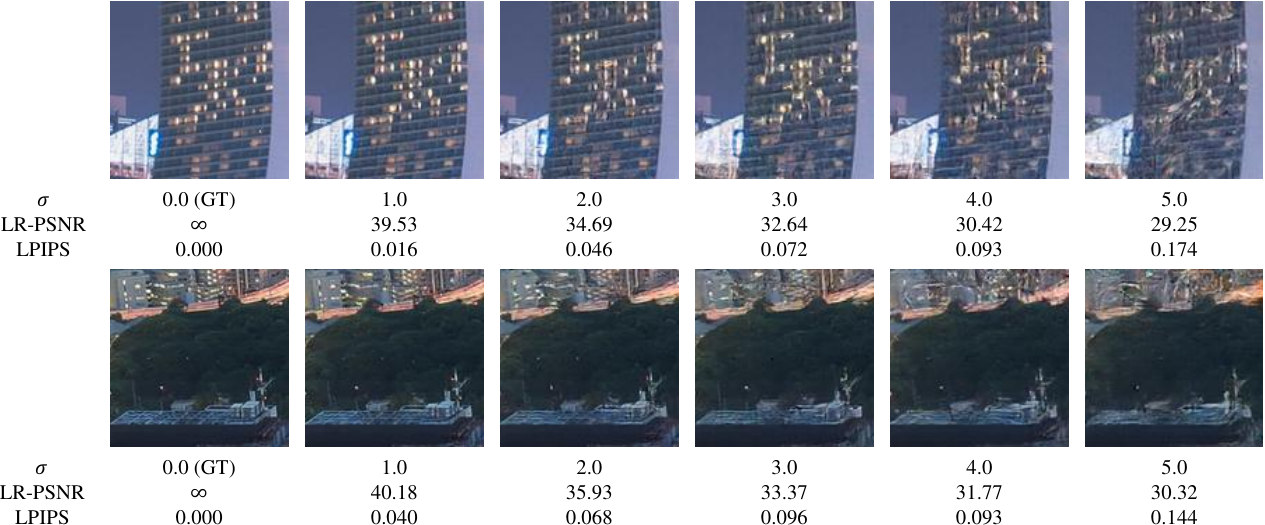}
    \caption{
        Visualization of fake samples generated by our proposed model with varying levels of distortion, controlled by $\sigma$.
        As $\sigma$ increases from 0.0 (the GT) to 5.0, the samples become progressively more distorted. 
        Notably, even at the most severe deformation, a certain level of LR-PSNR and LPIPS scores is maintained. 
        This demonstrates that the fake samples satisfy both photo-realism and LR correspondence, confirming they remain plausible (yet incorrect) solutions within the target manifold.
    }
    \label{fig:fake_samples}
\end{figure*}
Given varying noise levels, the trained model can synthesize fake samples with varying degrees of distortions, as demonstrated in \cref{fig:fake_samples}.

\begin{algorithm}[t]
\caption{Synthesizing Negative Samples for SISR model}
\label{alg:algorithm1}
\begin{algorithmic}[1]
\Require GT image $\mathbf{x}_{\text{GT}}$, noise level range $(\sigma_{\text{min}}, \sigma_{\text{max}})$, pre-trained fake generator $\psi(\cdot, \cdot)$, fidelity bias $\mathbf{x}_{\mu}$
\Ensure Fake samples $\{\dot{\mathbf{x}}^{i}\}_{i=1}^{N}$
\State $\mathbf{r} \leftarrow |\mathbf{x}_{\mu}-\mathbf{x}_{\text{GT}}|$
\For{$i=1$ \textbf{to} $N$}
    \State $\epsilon \sim \mathcal{N}(\mathbf{0}, \mathbf{I}), \quad \sigma \sim \mathcal{U}(\sigma_{\text{min}}, \sigma_{\text{max}})$
    \State $\mathbf{x}_{\sigma} \leftarrow \mathbf{x}_{\text{GT}} + \sigma \cdot \mathbf{r} \odot \epsilon$
    \State $\dot{\mathbf{x}}^{i} \leftarrow \psi(\mathbf{x}_{\sigma}, \sigma)$
\EndFor
\end{algorithmic}
\end{algorithm}

\subsection{Contrastive GANs for SISR}
\label{subsec:contrastive_gan}

Contrary to existing works using conventional GAN frameworks~\cite{ledig2017photo, wang2018esrgan, wang2021real}, we adopt a novel approach in which the discriminator plays a more complicated role.
Conventional discriminators in these studies learn to differentiate outputs of the generator from the GT image using two main strategies: 
(1) Estimating the probability that an input image is real~\cite{goodfellow2014generative}, or (2) predicting the probability that one image is more realistic than the other~\cite{jolicoeur2018relativistic}.
The first one, standard GAN (SGAN), formulates the adversarial loss for the generator as:
\begin{equation}
\label{eq:loss_sgan}
    \mathcal{L}_{\text{adv}}^{\text{SGAN}} = - \mathbb{E}_{\hat{\mathbf{x}}} \left[ 1 - \log  f_{D}^{s} \left( \hat{\mathbf{x}} \right) \right],
\end{equation}
where $\hat{\mathbf{x}} = f_{G}(\mathbf{x}_{\text{LR}})$.
On the other hand, relativistic average GAN (RAGAN) has the following form:
\begin{equation}
\label{eq:loss_ragan}
    \mathcal{L}_{\text{adv}}^{\text{RAGAN}} = -\mathbb{E}_{\mathbf{x}_{\text{GT}}} \left[ \log(1-f_{D}^{\text{RA}}(\mathbf{x}_{\text{GT}}, \hat{\mathbf{x}})) \right]
    - \mathbb{E}_{\hat{\mathbf{x}}} \left[ \log(f_{D}^{\text{RA}}(\hat{\mathbf{x}}, \mathbf{x}_{\text{GT}})) \right],
\end{equation}
where $f_{D}^{\text{RA}}(\mathbf{x}_{a}, \mathbf{x}_{b}) = \text{sigmoid} \left( f_C(\mathbf{x}_{a})-\mathbb{E}_{\mathbf{x}_{b}} \left[ f_{C}(\mathbf{x}_{b}) \right] \right)$ and $f_C(\cdot)$ is a critic that outputs an unnormalized logit.
Specifically, unlike the SGAN formulated in Eq. \ref{eq:loss_sgan}, RAGAN allows the generator to learn from enriched gradients calculated not only from generated samples but also from real samples.
Wang \etal~\cite{wang2018esrgan} demonstrates that this property improves sharper edges and more detailed textures of the generator's outputs.

Beyond the relativistic strategy, we propose a contrastive approach that utilizes the generated fake samples as unwelcome targets.
The first step is constructing the proper set of positive and negative samples for an SR prediction $\hat{\mathbf{x}} = f_G(\mathbf{x}_{\text{LR}})$.
From the SR network perspective—the generator in GAN literature—we aim to improve the network by teaching it not to generate undesirable outputs.
Thus, the adversarial loss for the generator should be formulated to repel the output from unwanted fakes (negatives) and attract it towards preferable samples (positives), whereas the discriminator does exactly the opposite thing to play a minimax game.

For each of batched GT images $\mathbf{x}_{\text{GT}} \in \mathbb{R}^{B \times 3 \times H \times W}$, we can synthesize $N$ random fake samples via our fake synthesizer.
However, if fake samples similar to the GT are regarded as negative, they can induce gradients in the direction of repelling the output from the GT.
Thus, we synthesize negative fake samples with high noise levels and call them \textit{off-manifold} solutions $\dot{\mathbf{x}}_{\text{off}} \in \mathbb{R}^{B \times N \times 3 \times H \times W}$, which indicates that they are far from the target space $\mathcal{M} \cap \mathcal{R}$.
Furthermore, the GTs can also be augmented to match the number of samples with the off-manifold fakes by synthesizing additional $N-1$ fakes with relatively low noise levels.
Since these augmented samples are located almost in the target space, we call them \textit{on-manifold} solutions $\dot{\mathbf{x}}_{\text{on}} \in \mathbb{R}^{B \times N \times 3 \times H \times W}$.

The $N$ on-manifold solutions derived from the $i$-th data, $\dot{\mathbf{x}}_{\text{on}}^{i} \in \mathbb{R}^{N \times 3 \times H \times W}$, are desirable targets for the SISR network's $i$-th output, $\hat{\mathbf{x}}^{i}$, as they simultaneously satisfy both LR correspondence and naturality.
We therefore define this set as the positive samples for the query $\hat{\mathbf{x}}^{i}$.
Theoretically, all other samples, including on-manifold solutions from different data $\dot{\mathbf{x}}_{\text{on}}^{j \neq i}$ and all off-manifold solutions $\dot{\mathbf{x}}_{\text{off}}$, should be treated as negatives. 
However, for computational efficiency, we simplify this by treating only the off-manifold solutions as negatives. 
This design choice is analyzed and validated in our ablation studies.


Consequently, for the generator, we establish the key samples for the $i$-th prediction as follows:
\begin{equation}
\label{eq:query_key_g}
    \mathbf{x}_{a}^{i} = \hat{\mathbf{x}}^{i}, \quad \mathbf{x}_{p}^{i} = \dot{\mathbf{x}}_{\text{on}}^{i} \quad \mathbf{x}_{n}^{i} =  \dot{\mathbf{x}}_{\text{off}}.
\end{equation}
For the discriminator, conversely, the roles of the two sets are exchanged:
\begin{equation}
\label{eq:query_key_d}
    \mathbf{x}_{a}^{i} = \hat{\mathbf{x}}^{i}, \quad \mathbf{x}_{p}^{i} = \dot{\mathbf{x}}_{\text{off}}^{i} \quad \mathbf{x}_{n}^{i} =  \dot{\mathbf{x}}_{\text{on}}.
\end{equation}
To take multiple positive samples into account, we use the supervised contrastive loss~\cite{khosla2020supervised} as the adversarial loss for both the generator and discriminator.
Before calculating it, the discriminator $f_{D}(\cdot)$ first maps the samples into vectors $\mathbf{z}_{a}^{i} \in \mathbb{R}^{D}$, $\mathbf{z}_{p}^{i} \in \mathbb{R}^{N \times D}$, and $\mathbf{z}_{n}^{i} \in \mathbb{R}^{(B \times N) \times D}$, respectively.
Then, the loss for the $i$-th sample is formulated as:
\begin{equation}
\label{eq:supcon}
    \mathcal{L}_{\text{supcon}}^{i} = \\
    -\frac{1}{N}\sum_{j} \log \frac{\exp(\langle\mathbf{z}_{a}^{i}, \mathbf{z}_{p}^{i, j}\rangle / \tau)}{\sum_{k} \exp(\langle\mathbf{z}_{a}^{i}, \mathbf{z}_{p}^{i, k}\rangle / \tau) + \sum_{l} \exp(\langle\mathbf{z}_{a}^{i}, \mathbf{z}_{n}^{i, l}\rangle / \tau)},
\end{equation}
where $\langle \cdot, \cdot \rangle$ denotes cosine similarity.

The temperature $\tau$ is a critical hyperparameter for the performance and stability of contrastive learning.
It is widely known that low temperature makes the model sensitive to hard negative samples, whereas high temperature improves training stability.~\cite{chen2020simple, khosla2020supervised}
To stabilize the minimax game, we minutely tune the temperatures for the generator and discriminator separately.
We mainly use $\tau_{G} = 0.06$ and $\tau_{D} = 0.15$ for the generator and discriminator, respectively, in the following experiments.

%% file: contents/experiments.tex
\section{Experiments}
\label{sec:experiments}

\subsection{Experimental Setups}

\subsubsection{Training Fake Sample Synthesizer}
For computational efficiency, our fake sample synthesizer $\psi(\cdot, \sigma)$ first divides the noisy input $\mathbf{x}_{\sigma} \in \mathbb{R}^{3 \times H \times W}$ into $4 \times 4$ non-overlapping patches.
The patchified input $\mathbf{x}_{\sigma}' \in \mathbb{R}^{48 \times H/4 \times W/4}$ is then fed into a U-Net with hidden dimensions set to $\left( 128, 256, 512 \right)$ for each stage.
A single self-attention layer is added after the two residual blocks at the deepest stage.
For GAN training, we employ the PatchGAN discriminator~\cite{isola2017image} as implemented by \cite{yu2024image}.

The training is conducted for 100k iterations with a batch size of 1024.
We utilize a cosine LR scheduler that warms up from $0$ to $10^{-4}$ for 5k iterations, and subsequently decreases to $10^{-5}$ until the end of training.
The coefficients in \cref{eq:loss_ae} are set as $\lambda_{\text{pix}} = 1.0$, $\lambda_{\text{LPIPS}} = 0.1$, and $\lambda_{\text{adv}} = 0.1$.
For stable GAN training, we use the hinge loss~\cite{lim2017geometric} and LeCAM regularization~\cite{tseng2021regularizing} with a decay factor of $0.999$.
All experiments were conducted with \texttt{bfloat16} precision using PyTorch~\cite{paszke2019pytorch}.

\subsubsection{Training MaCo-GAN}
To evaluate our proposed method, we employ the framework of AESOP~\cite{lee2025auto}, replacing only the adversarial losses for the SR network and the discriminator with our contrastive losses.
The weight of the adversarial loss for the SR network is also modulated from 0.005 to 0.2.
We use the DF2K dataset for training, which is a composite of DIV2K~\cite{agustsson2017ntire} and Flickr2K~\cite{lim2017enhanced}.
For computational efficiency and generalizability, training data is randomly cropped into $128 \times 128$ patches and augmented with random flips and rotations.

\subsubsection{Benchmark setups}
We assess our model on benchmark datasets including BSD100~\cite{martin2001database}, General100~\cite{dong2016accelerating}, Urban100~\cite{huang2015single}, Manga109~\cite{matsui2017sketch}, DIV2K-val~\cite{agustsson2017ntire}, and LSDIR-val~\cite{li2023lsdir}.
We report the PSNR and SSIM scores on the luminance channel for distortion metrics and LPIPS~\cite{zhang2018unreasonable}, DISTS~\cite{ding2020image}, and CLIPIQA~\cite{wang2023exploring} scores for perception metrics.
All evaluations are performed under $\times 4$ bicubic downsampling.

\subsection{Quantitative Results}

\begin{table}[t]
    \centering
    \scriptsize
    \caption{
        Comparison with existing SISR methods on various datasets for $\times 4$ upscaling.
        The best results of each group are highlighted in \textbf{bold}.
    }
    
    \begin{tabular}{l|c|cccc|c}
         \hline
         \multicolumn{2}{c|}{\multirow{2}{*}{\backslashbox[32mm]{Benchmarks}{Methods}}} & ESRGAN & SPSR & LDL & AESOP & \multirow{2}{*}{\textbf{Ours}} \\
         \multicolumn{2}{c|}{} & \cite{wang2018esrgan} & \cite{ma2021structure} & \cite{liang2022details} & \cite{lee2025auto} &  \\
         \hline \hline

         \multirow{4}{*}{BSD100} 
         & PSNR $\uparrow$       & 25.313    & 25.501    & 25.954    & 26.080   & \textbf{26.121} \\
         & SSIM $\uparrow$       & 0.6527    & 0.6596    & 0.6813    & 0.6841   & \textbf{0.6862} \\
         & LPIPS $\downarrow$    & 0.1616    & 0.1609    & 0.1535    & 0.1515   & \textbf{0.1505} \\
         & DISTS $\downarrow$    & 0.1165    & 0.1176    & 0.1163    & 0.1117   & \textbf{0.1097} \\
         \hline

         \multirow{4}{*}{General100} 
         & PSNR $\uparrow$       & 29.425    & 29.424    & 30.289    & \textbf{30.482} & 30.376 \\
         & SSIM $\uparrow$       & 0.8095    & 0.8091    & 0.8280    & \textbf{0.8335} & 0.8302 \\
         & LPIPS $\downarrow$    & 0.0879    & 0.0862    & 0.0796    & \textbf{0.0784} & 0.0786 \\
         & DISTS $\downarrow$    & 0.0874    & 0.0884    & 0.0801    & 0.0798 & \textbf{0.0782} \\
         \hline

         \multirow{4}{*}{Urban100} 
         & PSNR $\uparrow$       & 24.365    & 24.804    & 25.459    & 25.630  & \textbf{25.673} \\
         & SSIM $\uparrow$       & 0.7341    & 0.7474    & 0.7661    & 0.7724  & \textbf{0.7732} \\
         & LPIPS $\downarrow$    & 0.1229    & 0.1184    & 0.1084    & 0.1064  & \textbf{0.1045} \\
         & DISTS $\downarrow$    & 0.0880    & 0.0849    & 0.0793    & 0.0793  & \textbf{0.0772} \\
         \hline

         \multirow{4}{*}{Manga109} 
         & PSNR $\uparrow$       & 28.413    & 28.561    & 29.620    & \textbf{29.973} & 29.865 \\
         & SSIM $\uparrow$       & 0.8595    & 0.8590    & 0.8734    & \textbf{0.8827} & 0.8794 \\
         & LPIPS $\downarrow$    & 0.0649    & 0.0672    & 0.0544    & 0.0525 & \textbf{0.0511} \\
         & DISTS $\downarrow$    & 0.0471    & 0.0463    & 0.0355    & \textbf{0.0360} & 0.0362 \\
         \hline

         \multirow{4}{*}{DIV2K-val} 
         & PSNR $\uparrow$       & 28.175    & 28.182    & 28.819    & 29.079 & \textbf{29.114} \\
         & SSIM $\uparrow$       & 0.7759    & 0.7720    & 0.7897    & 0.7978 & \textbf{0.8019} \\
         & LPIPS $\downarrow$    & 0.1154    & 0.1099    & 0.0999    & 0.0977 & \textbf{0.0967} \\
         & DISTS $\downarrow$    & 0.0593    & 0.0546    & 0.0526    & 0.0518 & \textbf{0.0513} \\
         \hline

         \multirow{4}{*}{LSDIR-val} 
         & PSNR $\uparrow$       & 23.882    & 24.232    & 24.663    & 24.933 & \textbf{24.989} \\
         & SSIM $\uparrow$       & 0.6866    & 0.6966    & 0.7117    & 0.7220 & \textbf{0.7233} \\
         & LPIPS $\downarrow$    & 0.1378    & 0.1312    & 0.1180    & 0.1152 & \textbf{0.1146} \\
         & DISTS $\downarrow$    & 0.0764    & 0.0699    & 0.0650    & 0.0641 & \textbf{0.0632} \\
         \hline
    \end{tabular}
    
    \label{tab:comparison_reference}
\end{table}

We present the quantitative results in \cref{tab:comparison_reference}, comparing our method against several state-of-the-art GAN-based SISR models: ESRGAN~\cite{wang2018esrgan}, SPSR~\cite{ma2021structure}, LDL~\cite{liang2022details}, and our primary baseline, AESOP~\cite{lee2025auto}. 
The evaluation is conducted on six standard benchmark datasets (BSD100, General100, Urban100, Manga109, DIV2K-val, and LSDIR-val) using $\times 4$ upscaling. 

Our proposed method, which only replaces the adversarial loss of the baseline AESOP framework with ours, demonstrates a clear and consistent improvement in perceptual quality. 
Specifically, our model achieves better LPIPS and DISTS scores than AESOP on most datasets. 
This is achieved while maintaining highly comparable PSNR and SSIM scores, indicating an improved fidelity-perception trade-off.
It confirms the effectiveness and stability of our proposed contrastive adversarial objective.

\subsection{Qualitative Results}

\begin{figure}[t]
    \centering
    \includegraphics[width=1\linewidth]{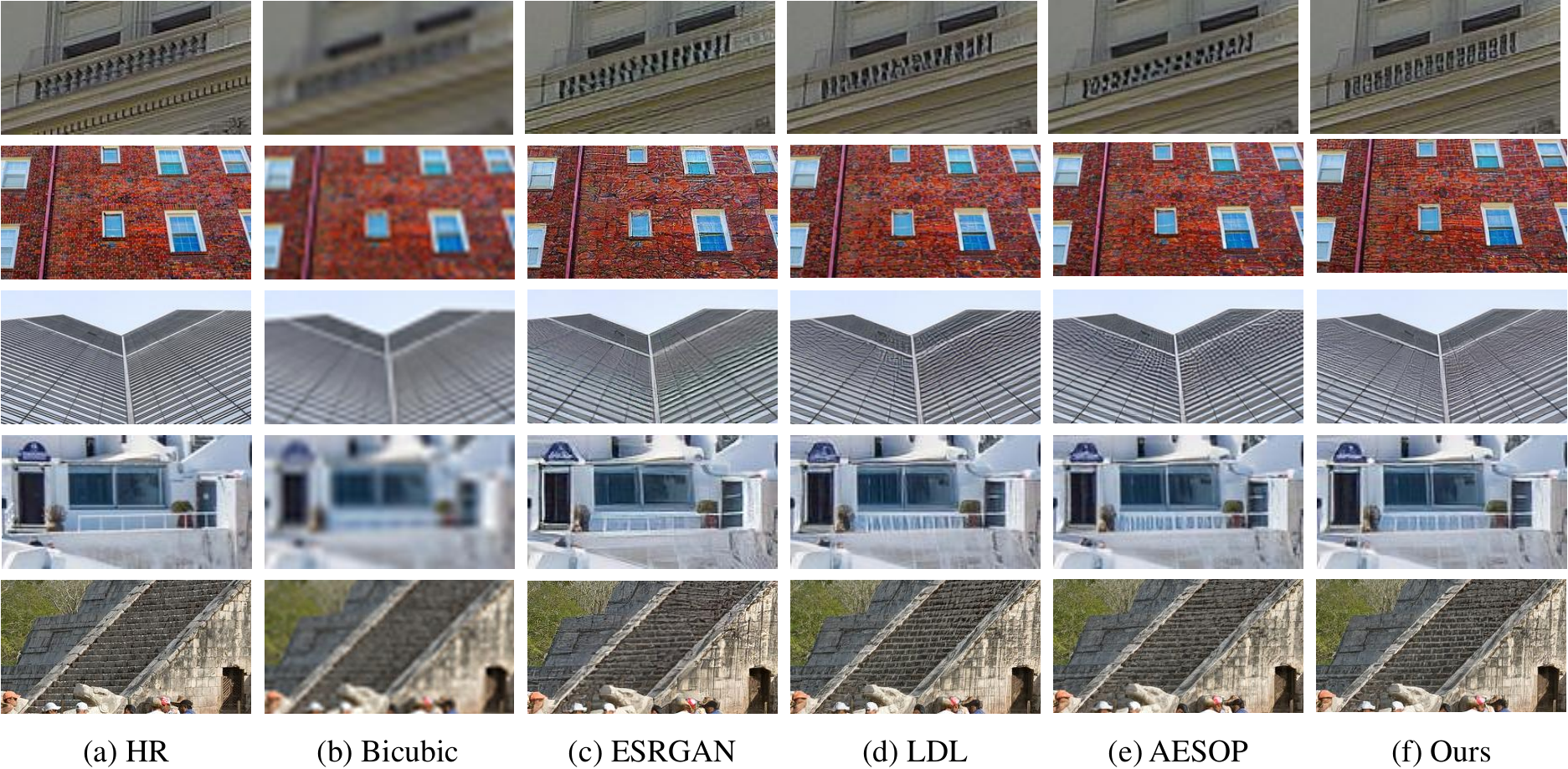}
    \caption{
        Visual comparison between our proposed method and previous models with the RRDB backbone~\cite{wang2018esrgan} for $\times 4$ upscaling.
        Zoom in for the best view.
    }
    \label{fig:visual_comparison}
\end{figure}

We present a qualitative comparison of our proposed method against several GAN-based SR models—ESRGAN~\cite{wang2018esrgan}, LDL~\cite{liang2022details}, and AESOP~\cite{lee2025auto}—on a selection of challenging images from standard benchmarks, as shown in \cref{fig:visual_comparison}.
While baseline models often exhibit noticeable artifacts in high-frequency structures, our proposed model reconstructs with higher fidelity and significantly fewer artifacts.

\subsection{Ablation Studies}
\subsubsection{Temperature of the contrastive losses}
To investigate the effect of temperature on the competition between the generator and discriminator, we conducted an ablation study. 
\cref{tab:ablation_temperature} shows the PSNR and LPIPS scores on the DIV2K validation set for each setting.
We first tested settings with identical temperatures ($\tau_G = \tau_D$), using values in $\{0.15, 0.10, 0.08, 0.06\}$.
In every case, the discriminator quickly overwhelms the generator and disrupts equilibrium early in training, resulting in suboptimal performance.
Thus, we opt to dampen the discriminator by increasing $\tau_D$, while fixing $\tau_G$.
This approach successfully regularizes the discriminator and prolongs the equilibrium, allowing the generator to reach a more optimal state with respect to the perception metric.

\begin{table}[t]
    \centering
    \scriptsize
    \caption{
        Ablation study on temperatures of adversarial losses for the generator and discriminator.
    }
    \begin{tabular}{c|cccc|ccc}
        \hline
        $\tau_G$ & 0.15 & 0.10 & 0.08 & 0.06 & 0.06 & 0.06 & 0.06 \\
        $\tau_D$ & 0.15 & 0.10 & 0.08 & 0.06 & 0.08 & 0.10 & 0.15 \\
        \hline \hline
        PSNR $\uparrow$ & \textbf{29.202} & 29.197  & 29.192 & 29.178 & 29.143 & 29.127 & 29.114 \\
        LPIPS $\downarrow$ & 0.1148 & 0.1127 & 0.1108 & 0.1047 & 0.0992 & 0.0973 & \textbf{0.0967} \\
        \hline
    \end{tabular}
    \label{tab:ablation_temperature}
\end{table}

\subsubsection{Number of key samples}
Prior studies in contrastive learning have shown a clear correlation between the number of key samples and representation learning performance~\cite{chen2020simple, khosla2020supervised}.
To validate how this key factor works in our framework, we conducted experiments with varying $N \in \{1, 2, 4, 8\}$.
Despite its advantage in representation learning, the results reported in \cref{tab:ablation_num_samples} indicate that the increased number of key samples hinders training stability instead.

This result, while seemingly counterintuitive, is theoretically sound and aligns with recent analyses of contrastive loss dynamics. 
Kim \etal~\cite{kim2025temperature} shows that increasing the number of keys, while fixing $\tau$, can make the contrastive loss function "too sharp," causing exploding gradients.
Thus, we can estimate that the increased number of keys causes the discriminator to encounter excessive gradients, which makes it easily overwhelm the SISR network.
This analysis validates our empirical finding: the contrastive minimax game is highly vulnerable to the discriminator's dominance.

\begin{table}[t]
    \centering
    \scriptsize
    \begin{minipage}{0.48\textwidth}
        \centering
        \caption{
            Ablation study on the number of fake samples synthesized from an anchor.
        }
        \resizebox{\textwidth}{!}{
        \begin{tabular}{c|cccc}
            \hline
            $N$ & 1 & 2 & 4 & 8 \\
            \hline \hline
            PSNR $\uparrow$ & 29.114 & 29.126 & 29.149 & \textbf{29.154} \\
            LPIPS $\downarrow$ & \textbf{0.0967} & 0.0978 & 0.0982 & 0.0994 \\
            \hline
        \end{tabular}
        }
        \label{tab:ablation_num_samples}
    \end{minipage}
    \hfill 
    \begin{minipage}{0.48\textwidth}
        \caption{
            Ablation study on the noise level boundary dividing on-manifold and off-manifold fakes.
        }
        \resizebox{\textwidth}{!}{
        \begin{tabular}{c|cccc}
            \hline
            $\gamma$ & 0.5 & 1.0 & 1.5 & 2.0 \\
            \hline \hline
            PSNR $\uparrow$ & \textbf{29.127} & 29.114 & 29.121 & 29.123 \\
            LPIPS $\downarrow$ & 0.0982 & \textbf{0.0967} & 0.0971 & 0.0973 \\
             \hline
        \end{tabular}
        }
        \label{tab:noise_threshold}
    \end{minipage}
\end{table}

\subsubsection{Range of noise levels}
We adopt a simple approach that sets the noise level range for on-manifold and off-manifold fake synthesis to $(0.0, \gamma)$ and $(\gamma, 5.0)$, respectively.
The boundary value $\gamma$ is a critical hyperparameter, as it directly controls the `realness' of the two synthetic distributions.
If $\gamma$ is set too high, most on-manifold solutions become overly distorted, guiding the generator's output to be unnatural. 
Conversely, if $\gamma$ is too low, most off-manifold solutions remain too realistic, providing an incorrect negative signal.
To find the optimal balance, we conducted an ablation study testing $\gamma \in \{0.5, 1.0, 1.5, 2.0\}$.
We found that $\gamma=1.0$ presented the most stable training and achieved the best fidelity and perceptual quality.

\begin{table}[t]
    \centering
    \scriptsize
    \caption{
        Ablation study on the key sample settings, each of which is denoted with the number of positives and negatives.
    }
    \begin{tabular}{c|ccc}
        \hline
        $\left(|P|, |N|\right)$ & $\left(NB, NB\right)$ & $\left(N, N(2B-1)\right)$ & $\left(N, NB\right)$ \\
        \hline \hline
        PSNR $\uparrow$     & \textbf{29.138} & 29.121 & 29.114 \\
        LPIPS $\downarrow$  & 0.0975 & 0.0981 & \textbf{0.0967} \\
         \hline
    \end{tabular}
    \label{tab:key_sample_definition}
\end{table}

\subsubsection{Definition of key samples}
In \cref{subsec:contrastive_gan}, in the perspective of the discriminator, we defined the off-manifold fakes aligned with $i$-th sample, $\dot{\mathbf{x}}_{\text{off}}^{i}$, as the positive samples for a $i$-th query $\hat{\mathbf{x}}^{i}$, and all on-manifold fakes, $\dot{\mathbf{x}}_{\text{on}}$, as the negative samples.
Note that we did not treat the unaligned off-manifold fakes $\dot{\mathbf{x}}_{\text{off}}^{j \neq i}$ as either positives or negatives.
If they are treated as positives, the supervised contrastive loss is calculated with $NB$ positives and $NB$ negatives.
Conversely, if they are assigned to negatives, the number of each key becomes $N$ and $N(2B-1)$, respectively.

We evaluated these alternatives and compared them to the baseline strategy with $B$ positives and $NB$ negatives.
In \cref{tab:key_sample_definition}, we can validate that taking account of the unaligned on-manifold fakes is not helpful in both cases.
The alternatives, despite the balance between the two key sets or the increased number of negative samples, performed worse than the baseline.
We hypothesize that this degradation results from the simplified task of the discriminator.

The first one with $NB$ positives and $NB$ negatives turns the discriminator's task into a simple binary classification that distinguishes between fake samples and real data, which is not much different from the conventional approaches.
In the second strategy with $N$ positives and $N(2B-1)$ negatives, assigning the unaligned off-manifold fakes to negatives is unprofitable, since they are regarded as negatives for both the generator and discriminator.
The $N(B-1)$ trivial terms can impede the discriminator from learning a meaningful representation by dominating the denominator of the contrastive loss.
Therefore, we construct the key sets only with aligned off-manifold fakes and all on-manifold fakes, yielding better perceptual quality at even lower cost.

\begin{table}[t]
    \centering
    \scriptsize
    \caption{
        Ablation study on the alternative methods for fake sample synthesis.
    }
    \begin{tabular}{c|ccccc}
        \hline
        Methods & Blur\&Sharp& Noise & Interp. & Learned  \\
        \hline \hline
        PSNR$\uparrow $   & 28.691  & 28.747  & 28.962   & \textbf{29.114}   \\
        LPIPS$\downarrow$ & 0.1176  & 0.1187  & 0.1023   & \textbf{0.0967}   \\
        \hline
    \end{tabular}
    \label{tab:ablation_fake_synthesis}
\end{table}

\subsubsection{Data synthesis strategies}
To demonstrate the necessity of our proposed fake synthesizer, we compare it against three training-free alternative strategies for generating fake samples. 
These alternatives are computationally inexpensive, preserve LR correspondence, and allow the degree of distortion to be easily modulated. 
The first approach, inspired by Wu \etal~\cite{wu2023practical}, constructs on-manifold and off-manifold samples by sharpening and blurring the ground truth (GT) with random parameters, respectively. 
The second approach directly utilizes the raw noisy input, $\mathbf{x}_{\sigma}$, which is equivalent to sampling from an anisotropic Gaussian distribution $\mathcal{N}\left(\mathbf{x}_{\text{GT}}, \sigma^2 \cdot \text{diag} (\mathbf{r}) \right)$. 
A critical limitation of both methods is that they yield trivial off-manifold samples—either excessively blurry or noisy—that lie exceptionally far from the natural image manifold. 
Consequently, they reduce the discriminator's objective to a simple separation task, allowing it to quickly overpower the generator. 
To mitigate this, the third strategy employs a simple linear interpolation between the GT and the generator's current prediction:
\begin{equation}
    \dot{\mathbf{x}}^{i} = \alpha \cdot \mathbf{x}_{\text{GT}} + (1-\alpha) \cdot \hat{\mathbf{x}},
\end{equation}
where $\alpha \sim \mathcal{U}(0, 1)$.

We evaluate these alternatives against our proposed synthesizer, with the results reported in \cref{tab:ablation_fake_synthesis}. 
As anticipated, the two naive approaches, blur/sharpen and raw noise, lead to significant degradations across both distortion and perceptual metrics. 
The interpolation strategy achieves highly competitive performance, as it consistently challenges the discriminator with off-manifold samples that are perceptually superior to the current SR output. 
Nevertheless, it still falls short of our proposed method. 
We therefore conclude that our learned fake sample synthesizer—which dynamically generates challenging fakes that are both LR-consistent and perceptually plausible—is indispensable to the success of our contrastive adversarial framework.

%% file: contents/conclusion.tex
\section{Conclusion}

In this paper, we presented MaCo-GAN, a novel contrastive generative adversarial framework that addresses the limitations of conventional, unconditional discriminators in perceptual SISR. 
By replacing the standard adversarial objective with a supervised contrastive loss, our method enforces strict conditional realism. 
The core of this framework is the fake sample synthesizer, which dynamically generates a spectrum of perceptually plausible and LR-consistent fake samples. 
By utilizing these synthesized samples to construct informative `on-manifold' and `off-manifold' sets, we establish a robust contrastive minimax game that guides the generator's predictions toward the optimal target manifold. 
Extensive experiments across multiple benchmarks demonstrate that MaCo-GAN consistently achieves a superior perception-distortion trade-off compared to state-of-the-art baselines, without incurring prohibitive computational costs.

Moving forward, our contrastive strategy opens several promising avenues for future research.
While our current fake sample synthesizer operates effectively under fixed degradation assumptions, extending this mechanism to model complex, spatially variant degradations for real-world blind SISR is a compelling next step. 
Furthermore, adapting this conditional contrastive framework to other generative architectures, such as diffusion models, or applying it to broader ill-posed image restoration tasks (e.g., deblurring and inpainting), presents an exciting direction for future exploration.

%% file: contents/supplementary.tex
\clearpage
\setcounter{page}{1}
\setcounter{section}{0}
\renewcommand{\thesection}{\Alph{section}}

\title{Supplementary Material for \\ MaCo-GAN: Manifold-Contrastive Adversarial Learning for Single Image Super-Resolution} 

\titlerunning{MaCo-GAN: Supplementary Material}

\author{Daeyoung Han\inst{1}\orcidlink{0000-0003-0368-8675} \and
Seongmin Hwang\inst{2}\orcidlink{0000-0003-3313-8586} \and
Moongu Jeon\inst{1}\orcidlink{0000-0002-2775-7789}}

\authorrunning{D. Han, et al.}

\institute{
Department of Electrical Engineering and Computer Science, Gwangju Institute of Science and Technology, Gwangju, Republic of Korea \\
\email{\{xesta120, mgjeon\}@gist.ac.kr} \and
Department of AI Convergence, Gwangju Institute of Science and Technology, Gwangju, Republic of Korea \\
\email{sm.hwang@gm.gist.ac.kr}
}

\maketitle

\section{Analysis}
\label{sec:supp_analysis}

\subsection{Evaluation of the Fake Sample Synthesizer}

\begin{figure}[t]
    \centering
    \begin{minipage}[t]{0.48\linewidth}
        \centering
        \includegraphics[width=1.0\linewidth]{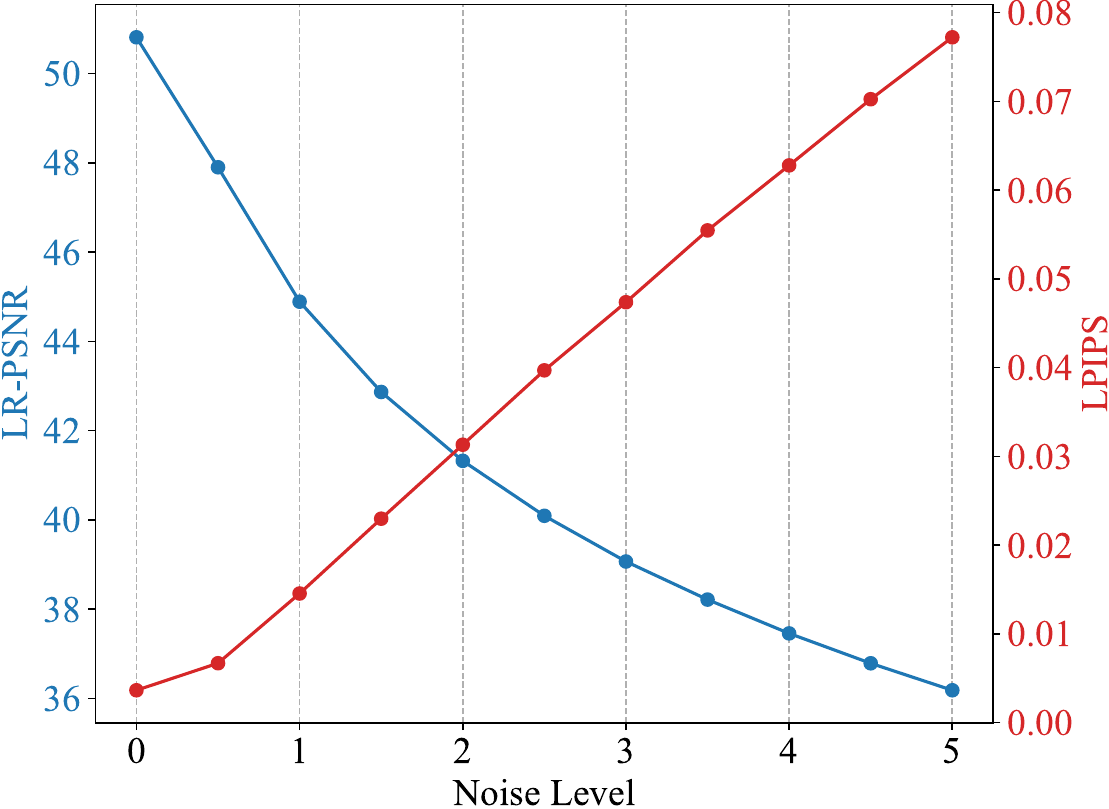}
        \caption{
            Quantitative analysis of our synthesized fakes.
            The applied noise level ($\sigma$) shows a significant correlation with the perceptual distance from the GT (LPIPS, red), validating its effectiveness in modulating the difficulty of the outputs while maintaining high LR correspondence (LR-PSNR, blue).
        }
        \label{fig:fake_scores}
    \end{minipage}
    \hfill
    \begin{minipage}[t]{0.48\linewidth}
        \centering
        \includegraphics[width=0.7\linewidth]{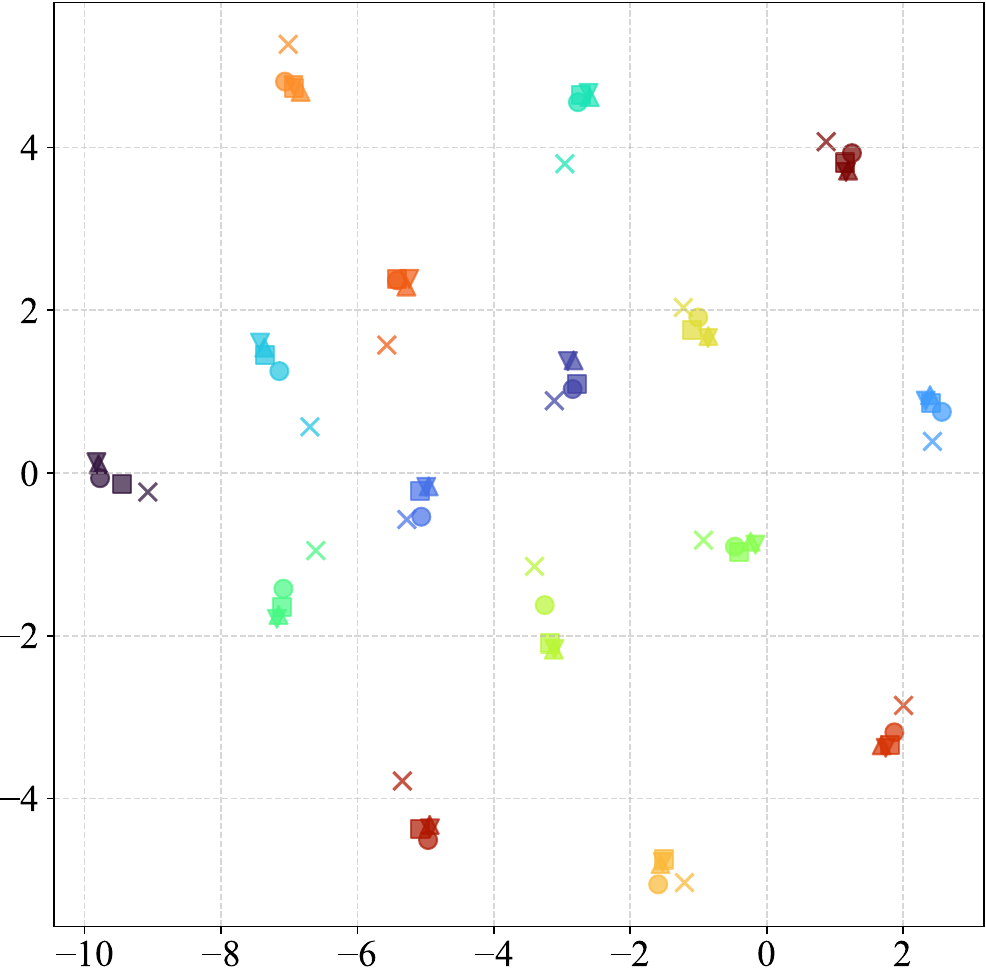}
        \caption{
            t-SNE visualization of the discriminator’s feature embeddings extracted from GTs ($\bullet$), on-manifold fakes ($\blacktriangle$), off-manifold fakes ($\blacktriangledown$), blurry LR (\texttimes), and SR predictions ($\blacksquare$). 
            Samples with the same color share a common LR image.
        }
        \label{fig:tsne_contrastive_features}
    \end{minipage}
\end{figure}

To validate our fake sample synthesizer, we performed an evaluation on the DIV2K validation set. 
We generated 1,000 samples from randomly cropped HR patches for each noise level $\sigma$ in the range $\{0.0, 0.5, \dots, 5.0\}$. The resulting average LR-PSNR and LPIPS scores are plotted in \cref{fig:fake_scores}. 
The results verify that $\sigma$ effectively controls perceptual distortion; as $\sigma$ increases, the LPIPS score steadily rises while the LR-PSNR score decreases. 
Critically, the LR-PSNR remains high across the entire range, confirming that low-resolution correspondence is well-preserved even under significant degradation.

Notably, even our most distorted fakes at $\sigma=5.0$ exhibit a perceptual quality (LPIPS $\approx 0.08$) that remains competitive with current state-of-the-art models, which typically score around 0.09. 
We specifically cap the maximum noise level at $\sigma_{\max} = 5.0$ in a data-driven manner to ensure the generated samples act as hard negatives. 
Pushing the noise level beyond this threshold yields excessively distorted fakes that become trivial for the discriminator to separate, thereby neutralizing their training value. 
By aligning this upper bound with the actual error space of current SOTA predictions, we guarantee that the synthesizer yields a broad spectrum of highly challenging, perceptually plausible fakes. 
These calibrated targets force the network to learn fine-grained, high-frequency differences from the GT rather than relying on rudimentary separation cues. 
We note that the synthesizer sacrifices perfect autoencoding capacity at $\sigma=0$, as evidenced by the imperfect scores at the origin, to achieve this robust generative performance for all $\sigma > 0$.

\subsection{Representation Space of the Discriminator}
We explored the representation space of the learned discriminator to analyze the dynamics of the contrastive GAN.
We present a t-SNE visualization of the discriminator's embeddings in \cref{fig:tsne_contrastive_features}. 
The plot first confirms that training proceeds effectively without mode collapse.
Additionally, we observe that the primary organizing principle for the embeddings is not the sample type—GT, SR, or Fake—but rather the shared visual content.
In other words, clusters form around samples corresponding to a common LR image. 
Notably, the representations of the synthesized fake samples are much more similar to the GTs compared to the LRs upscaled to the same resolution.
The average cosine similarity between LRs and GTs measures $0.7636$, while those of on-manifold fakes, off-manifold fakes, and SR prediction measure $0.9248$, $0.9206$, and $0.9439$, respectively.

This concentration is reasonable and expected because, in \cref{eq:query_key_g} and \cref{eq:query_key_d}, the majority of negative samples are unaligned with the query.
By dividing the negative samples $\mathbf{z}_{n}^{i}$ into aligned ($\mathbf{z}_{n+}^{i}$) and unaligned ($\mathbf{z}_{n-}^{i}$) subsets, the supervised contrastive loss in \cref{eq:supcon} can be expressed as:
\begin{equation}
\label{eq:supcon2}
    \mathcal{L}_{\text{supcon}}^{i} =
    -\frac{1}{N} \sum_j \log \frac{{s}_{p}^{i, j}}{\sum_k {s}_{p}^{i, k} + \sum_l s_{n+}^{i, l} + \sum_m s_{n-}^{i, m}},
\end{equation}
where $s_c^{i, j} = \exp \left( \langle \mathbf{z}_{a}^{i}, \mathbf{z}_{c}^{j} \rangle / \tau \right)$ for $c \in \{p,n^+,n^-\}$.
Crucially, most unaligned negatives for $\mathbf{z}_{a}^{i}$ are also unaligned with respect to any other query.
Specifically, each similarity term $s_{n-}^{i, m}$ for $i \in [1, B]$ and $m \in [1, B\cdot N] \setminus [(i-1)\cdot N+1, i \cdot N]$ appears $(B-1)$ times in the denominator of \cref{eq:supcon2} when calculating the total loss $\sum_i \mathcal{L}_{\text{supcon}}^{i}$.
Therefore, minimizing the supervised contrastive loss prioritizes reducing the similarity between a query and its unaligned negative samples.
This drives the discriminator to spread the representation over the entire embedding space $S^{D-1} = \{ \mathbf{z} \in \mathbb{R}^{D} \mid \|\mathbf{z}\|_2 = 1 \}$.

Such a diffuse representation space in the discriminator is particularly beneficial for improving the generator and aligns with our design intent.
When gathered into a local cluster, a query embedding and its aligned positives and negatives exhibit cosine similarities close to 1.
Consequently, $\mathbf{z}_{p}^{i}$ and $\mathbf{z}_{n+}^{i}$ serve as easy positive and extremely hard negative samples for the query $\mathbf{z}_{a}^i$, respectively.
As proven by Khosla \etal~\cite{khosla2020supervised}, these samples have marginal gradient contributions with respect to the query embedding before normalization.
In contrast, the unaligned negatives $\mathbf{z}_{n-}^{i}$ show cosine similarities near zero with $\mathbf{z}_{p}^{i}$.
Therefore, the SR network is primarily optimized by receiving a signal from the discriminator that penalizes predictions resembling off-manifold solutions.

\section{Additional Results}

\subsection{Computational Complexity}

\begin{table}[t]
    \centering
    \scriptsize
    \caption{
        Comparison of computational training overhead. 
    }
    \begin{tabular}{c|cccc|c}
         \hline
         \multirow{2}{*}{Methods} & SRGAN & ESRGAN & LDL & AESOP & \multirow{2}{*}{\textbf{Ours}} \\
          & \cite{ledig2017photo} & \cite{wang2018esrgan} & \cite{liang2022details} & \cite{lee2025auto} & \\
         \hline \hline
         GPU Memory (GB)        & 12.81 & 12.81 & 12.86 & 20.38 & 21.03 \\
         Throughput (iter/s)    & 1.33 & 1.30  & 1.16  & 0.82  & 0.77 \\
         \hline
    \end{tabular}
    \label{tab:computational_complexity}
\end{table}

To validate the efficiency of our proposed framework, we compare the computational complexity of MaCo-GAN against existing state-of-the-art models. 
The peak GPU memory consumption and training throughput (iterations per second) are evaluated under identical settings: A mini-batch of 256 cropped patches with the size of $128 \times 128$, distributed on 4 NVIDIA RTX A5000.
As reported in \cref{tab:computational_complexity}, integrating the fake sample synthesizer and our manifold-contrastive objective introduces only a marginal overhead compared to our primary baseline, AESOP~\cite{lee2025auto}. 
Specifically, our method incurs a minor increase in memory of 0.65 GB and a negligible reduction in training throughput of 0.05 iterations per second.
Given that prior contrastive SISR approaches often require computationally heavy auxiliary networks that increase memory consumption, our framework offers a favorable trade-off, achieving significant perceptual improvements with minimal additional computational cost.

\subsection{Quantitative Results}
\begin{table}[t]
    \scriptsize
    \centering
    \caption{
        Comparison with existing SISR methods on HR datasets for $\times 4$ upscaling.
        The best results of each group are highlighted in \textbf{bold}.
    }
    \begin{tabular}{c|c|ccc|c}
        \hline
        \multicolumn{2}{c|}{\multirow{2}{*}{\backslashbox[32mm]{Benchmarks}{Methods}}} & ESRGAN & LDL & AESOP & \multirow{2}{*}{\textbf{Ours}} \\
        \multicolumn{2}{c|}{} & \cite{wang2018esrgan} & \cite{liang2022details} & \cite{lee2025auto} & \\
        \hline \hline
        \multirow{4}{*}{DIV2K-val}  & NIQE $\downarrow$     & 3.0489    & 2.9683    & 3.0227    & \textbf{2.9389}    \\
                                    & MANIQA $\uparrow$     & 0.3812    & 0.3661    & 0.3588    & \textbf{0.3826}    \\
                                    & MUSIQ $\uparrow$      & 64.619    & 64.586    & 64.056    & \textbf{65.448}    \\
                                    & CLIP-IQA $\uparrow$   & \textbf{0.6278}    & 0.6004    & 0.6034    & 0.6042    \\
        \hline
        \multirow{4}{*}{LSDIR-val}  & NIQE $\downarrow$     & 3.0114    & 2.6812    & 2.6633   & \textbf{2.6600} \\
                                    & MANIQA $\uparrow$     & \textbf{0.4932}    & 0.4634    & 0.4592   & 0.4812 \\
                                    & MUSIQ $\uparrow$      & 71.438    & 71.371    & 71.339   & \textbf{71.991} \\
                                    & CLIP-IQA $\uparrow$   & \textbf{0.6824}    & 0.6594    & 0.6641   & 0.6699 \\
         \hline
    \end{tabular}
    \label{tab:comparison_nonreference}
\end{table}

We evaluated our method against the previous SOTA models by measuring non-referential metrics for image quality assessment (IQA), NIQE~\cite{mittal2012making}, MANIQA~\cite{yang2022maniqa}, MUSIQ~\cite{ke2021musiq}, and CLIP-IQA~\cite{wang2023exploring}.
To compare the naturality of outputs, the evaluation is conducted on two high-resolution natural datasets, DIV2K~\cite{agustsson2017ntire} and LSDIR~\cite{li2023lsdir}.
The scores are reported in \cref{tab:comparison_nonreference}, which shows that our method performs best in most cases.

\subsection{Qualitative Results}

\begin{figure}[t]
    \centering
    \includegraphics[width=1\linewidth]{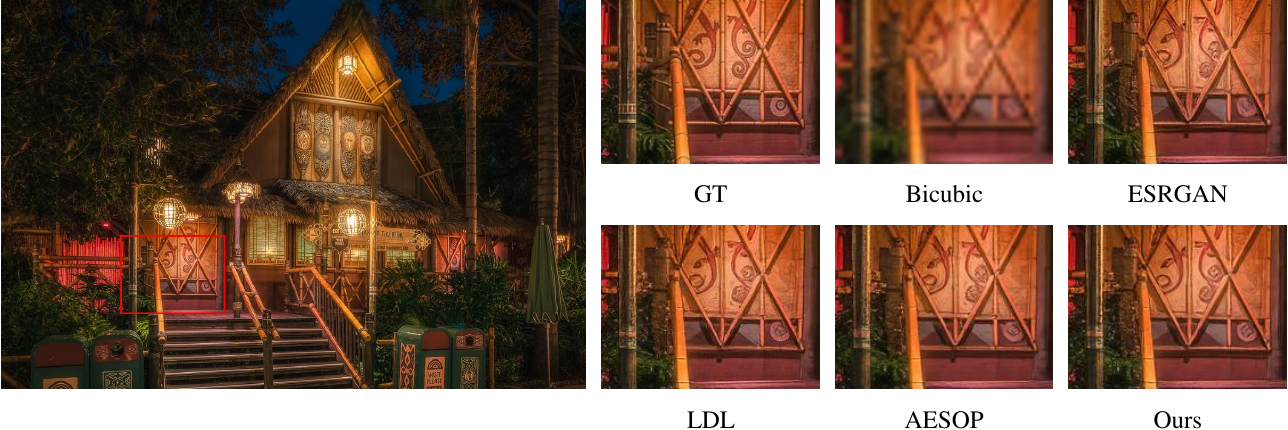}
    \vfill
    \includegraphics[width=1\linewidth]{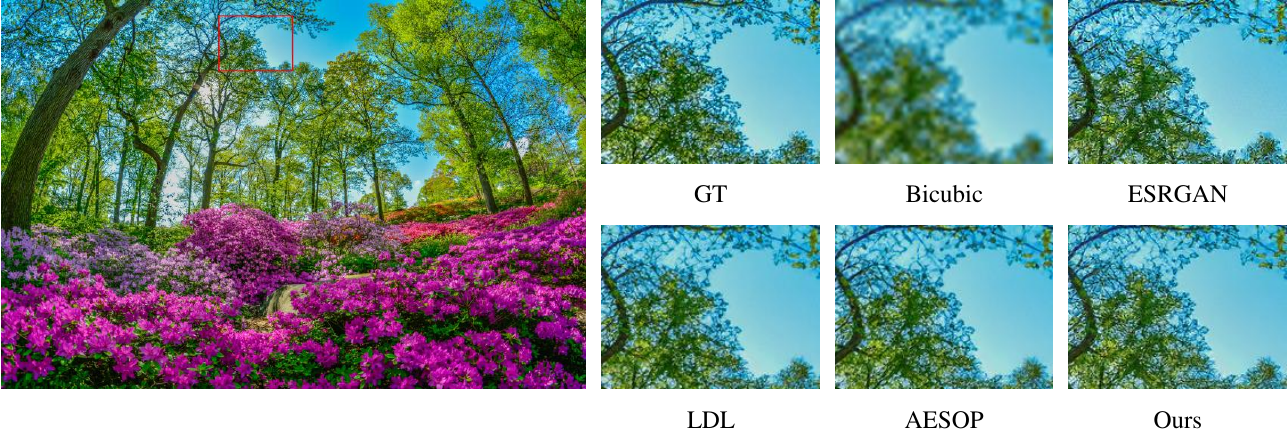}
    \vfill
    \includegraphics[width=1\linewidth]{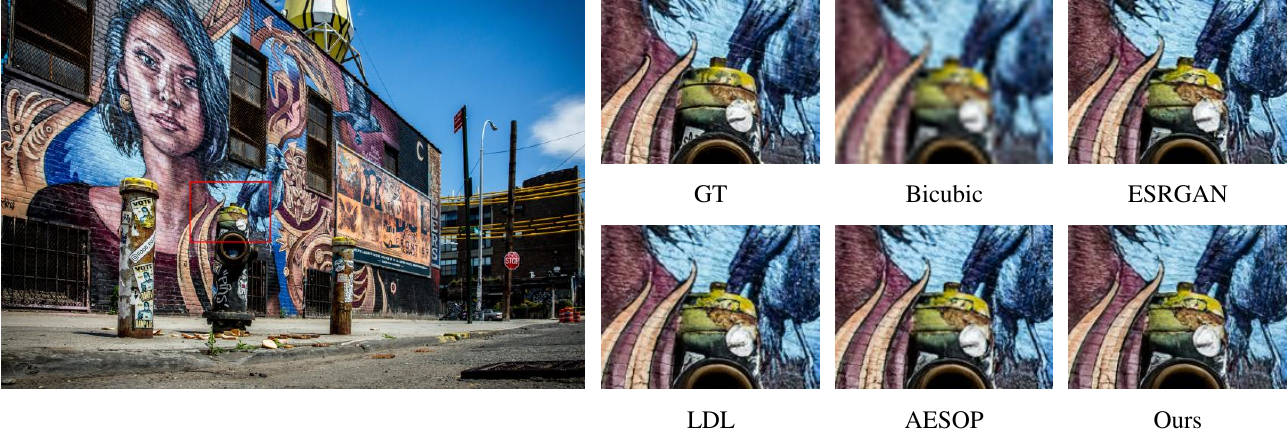}
    \caption{
        Visualization of our proposed model and the baseline methods for the bicubic ×4 SR task. 
        Zoom in for the best view.
    }
    \label{fig:ablation_quantitative_comparison}
\end{figure}

In this section, we present additional visual results of our method, comparing with ESRGAN~\cite{wang2018esrgan}, LDL~\cite{liang2022details}, and AESOP~\cite{lee2025auto}.
\cref{fig:ablation_quantitative_comparison} demonstrates that our proposed model is superior and robust, especially in reconstructing fine structural details.